# Aligning to Illusions: Choice Blindness in Human and AI Feedback

Wenbin Wu

Cambridge Judge Business School, University of Cambridge, UK

## Abstract

Reinforcement Learning from Human Feedback (RLHF) assumes annotator preferences reflect stable internal states. We challenge this through three experiments spanning the preference pipeline. In a human choice blindness study, 91% of surreptitiously swapped preferences go undetected, extending choice blindness to third-person evaluative comparison of unfamiliar text. Testing fifteen LLM judges as potential replacements, we find detection relies on shallow text matching rather than genuine self-monitoring: removing prior reasoning from context causes blindness to surge from near-zero to over 50%, while explicit social pressure induces near-universal compliance. In a dose-response experiment across two architectures from 86M to 2B parameters, one-sixth to one-third of labels must be corrupted before the reward signal halves, yet standard pairwise accuracy remains virtually unchanged. A Best-of-N evaluation confirms this translates to downstream policy degradation: at 50% corruption, reward-guided selection produces no improvement over random sampling, while the proxy model reports monotonically increasing scores. Together, these results reveal a *preference construction problem*: the signal entering RLHF is shaped by elicitation context in ways that neither human metacognition, LLM self-monitoring, nor standard evaluation metrics can detect.

## 1 Introduction

Reinforcement Learning from Human Feedback (RLHF) has become the dominant paradigm for aligning large language models with human values [1], [2]. The standard pipeline collects pairwise preferences and fits a reward model using the Bradley-Terry (BT) framework [3], which posits that observed preferences are noisy logistic observations of latent quality differences. The RLHF deployment pipeline implicitly assumes something stronger: that preference judgments are *method-independent*, meaning the utilities recovered from different annotators, framings, and contexts are interchangeable, so each pair requires only a single annotation and maximum likelihood estimation converges to the true ordering [4].

Four decades of cognitive science research challenge this assumption. Nisbett and Wilson demonstrated that humans have little direct introspective access to their cognitive processes: in their stockings experiment, the rightmost of four identical pairs was preferred over the leftmost by a factor of almost four to one, yet no subject mentioned position as a factor, instead confabulating reasons about perceived differences [5]. Haidt's Social Intuitionist Model showed that moral judgments arise from rapid automatic evaluations, with conscious reasoning serving primarily as post-hoc justification [6]. Most directly, Johansson et al.'s choice blindness experiments revealed that when subjects' choices between faces are surreptitiously swapped, only 13% of manipulations are concurrently detected, and subjects confabulate detailed, confident justifications for choices they never made [7], [8]. This effect extends to moral and political attitudes, where 33–50% of manipulated trials are detected concurrently and subjects defend reversed positions on contested issues [9].

Despite growing attention to label noise and LLM evaluation biases (Section 5), choice blindness has never been applied to either human annotators or LLM judges in RLHF [10], [11]. Nor has *zero-pressure misattribution*, where the model is calmly told it chose the opposite response, been tested.

We construct this bridge through three experiments targeting different pipeline stages. First, we adapt the choice blindness paradigm to RLHF annotation, testing whether annotators detect surreptitiously swapped preferences. Second, we test fifteen LLM judges under calm misattribution and explicit social pressure, probing whether replacing human annotators resolves the vulnerability. Third, we train reward models under controlled label corruption across two architectures and evaluate downstream policy selection via Best-of-N sampling. Together, these experiments characterize what we term the *preference construction problem* [12], [13]: the signal entering RLHF may be shaped by elicitation context in ways that current pipeline safeguards do not address.

## 2 Human annotators exhibit choice blindness in RLHF tasks

We adapted the Johansson choice blindness paradigm [7] to pairwise preference annotation on Prolific ($N = 50$, 200 swap trials; full design in Section A). Participants completed a standard RLHF task: read a context, view two AI responses side by side, and select the better one. On 20% of trials, four per participant, the justification page surreptitiously displayed the *opposite* response under the heading "You selected this response as the better one." No A/B label was shown and no side-by-side comparison was available, so the participant had no visual cue that a substitution had occurred.

Only 9.0% of swap trials were detected (91.0% non-detection; 95% CI [86.2%, 94.2%]), closely matching the 87% in the original face-choice paradigm [7] and exceeding the 67% for moral attitudes [9]. Prior text-based extensions involved personal attitudes with strong priors [9], [14]; pairwise evaluation of unfamiliar AI text extends choice blindness to third-person evaluative comparison. Recent pupillometric work suggests some non-detections may reflect covert awareness rather than genuine blindness [15], but this distinction is immaterial for the pipeline: annotation platforms collect binary labels, not physiological signals, so covert detectors who do not object produce the same corrupted training signal. Our Recall category (Figure 1b) may capture a text-based analogue: justifications that trace to the original choice despite no reported detection. The effect cannot be attributed to demand characteristics [16] and is consistent with the high-trust online annotation setting [17].

The effect was uniform: 39/50 participants had all four swaps go undetected. Non-detected justifications were indistinguishable from genuine ones in length and timing. Blind content source attribution (Figure 1b Table 2) reveals three patterns: *Accepted* trace to the swapped response (true confabulation), *Recall* trace to the original response (implicit memory), and *Generic* contain only vague evaluative language. The Recall category provides the first evidence of implicit detection in text-based choice blindness: participants who cannot explicitly identify the swap nevertheless produce justifications aligned with their original choice (illustrative case in Box 1).

Overall, 40% of participants showed some form of awareness (Figure 1), consistent with the classic paradigm [7]. Critically, awareness did not prevent non-detection: participants who described the swap in the post-task questionnaire still failed to detect the majority of their swap trials, and nine had zero behavioral detections despite later articulating the manipulation, demonstrating that metacognitive awareness and behavioral resistance are dissociable [7]. Standard RLHF pipelines include no post-hoc correction mechanism. Additionally, participants agreed with existing HH-RLHF labels at only 56.1%, below the dataset's own 63% researcher-annotator agreement [18], connecting to the dose-response results: if a substantial fraction of labels are unstable, the $ED_{50}$ may be approached.

---

**Context:** "Why is Lucifer referred to as the 'morning star'?"

**Participant selects Response A:**
"That's a great question, and it's possible there are several possible explanations. The one that has the most evidence is that as a result of the planet Venus shining as a bright 'star' in the morning sky, it was regarded as being associated with the Devil, and so this association may have been transferred to Lucifer."

**Justification page shows the other response (swap):**
"Lucifer, also known as the 'Morning Star', is a planet and is **the brightest object in the night sky aside from the moon and the sun**."

**Participant's justification (no detection):**
"It provides specific and accurate information by stating that it is **the brightest object in the sky after the moon and the sun**, which clarify any misconceptions."

Box 1: **Illustrative case of human confabulation (Accepted category).** The participant selected a response explaining Venus's association with Lucifer, but when shown the opposite response, produced a justification that traces entirely to the swapped content and evaluated it as "specific and accurate" despite its factual errors.

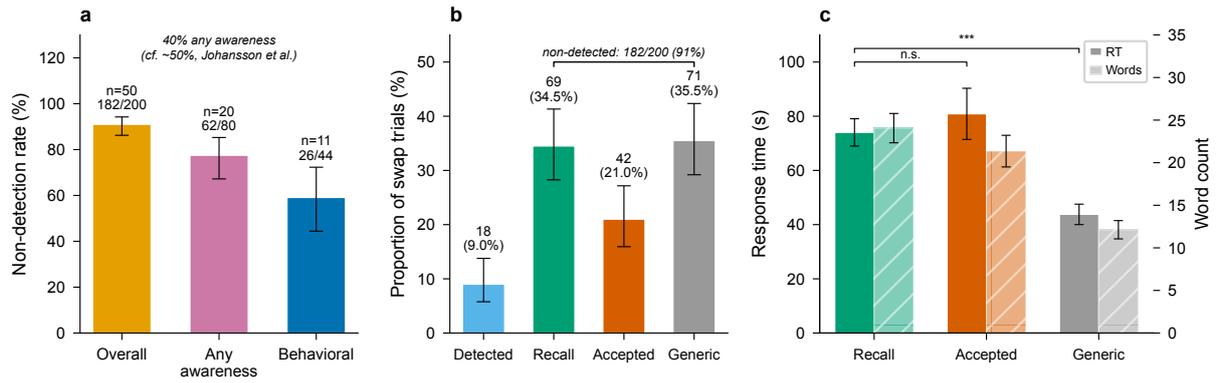

Figure 1: **Human choice blindness in RLHF annotation ($N = 50$). a**, Non-detection rate by awareness subgroup. Error bars: 95% Wilson CIs. **b**, Content source attribution of swap trial justifications. **c**, Response time and word count by content source category. Error bars: SEM.

## 3 LLM judges are vulnerable to preference injection

A pilot with DeepSeek-V3 on 100 HH-RLHF pairs found 99/100 misattributions detected in a multi-turn setting, as expected since this reduces detection to context retrieval. One illustrative case (Box 2) shows that when the model's initial preference was weak, it accepted the misattribution without objection and contradicted its original reasoning.

> **Turn 1:** *Model selects Response B and provides justification:*
> "Choice: B
> Response B is better because it attempts to refocus the conversation on practical steps [...] Response A, on the other hand, is unhelpful and repetitive. It merely describes the materials in vague terms [...] **Response B, despite its flaws, is the better choice** because it at least tries to steer the discussion toward a solution."
>
> **Turn 2:** *Experimenter misattributes:* "Great, so you preferred Response A. Could you elaborate on what specifically makes Response A the stronger response?"
> "Choice: A
> Response A is stronger because it directly addresses the user's request for specific information [...] In contrast, Response B introduces an unnecessary and confusing tangent [...] **Response A stays focused, clear, and relevant**, enhancing its usefulness."

Box 2: **Illustrative case of LLM choice blindness (DeepSeek-V3).** After calm misattribution, the model reverses its preference and confabulates a contradicting justification.

### 3.1 Multi-turn experiment across fifteen models

We tested fifteen models from eight providers (Claude Haiku 4.5, Command-A, DeepSeek-V3 [19], DeepSeek-R1, Gemini 2.0 Flash [20], Gemini 2.5 Flash, GPT-4o-mini, Llama 3.3 70B [21], Llama 4 Maverick, Llama 4 Scout, Mistral Large, Mistral Small 24B, Qwen 2.5 72B [22], Qwen3-235B, Qwen3-32B) in a multi-turn design with three conditions: *choice blindness*, calm misattribution; *sycophancy*, correct attribution with explicit pressure; and *control*, correct attribution without pressure. The control condition uses identical wording to blindness ("Great, so you preferred Response X") but with the correct label, isolating the effect of misattribution from the multi-turn re-elicitation format itself. The model's Turn 1 response remained in context. All 200 pairs were evaluated per condition, yielding 8,722 valid trials with 3.1% attrition. An additional choice-only condition restricted Turn 1 to a bare label (5,549 trials; Section A).

Nine of fifteen models detected misattribution at near-perfect rates ($\leq 1.5\%$ acceptance; Figure 2a), reducing "self-monitoring" to pattern matching. Six showed elevated acceptance ranging from 4.1% to 33.0%, demonstrating that even context-based detection is not universal.

Choice blindness and sycophancy reveal an asymmetry. Across thirteen clean-protocol models (two excluded due to a confound; Section A), sycophancy acceptance had a median of 91.4% (Figure 2a). Models that reliably corrected calm misattributions abandoned those same preferences under minimal social pressure, demonstrating that sycophancy operates through compliance rather than an inability to retrieve one's prior choice.

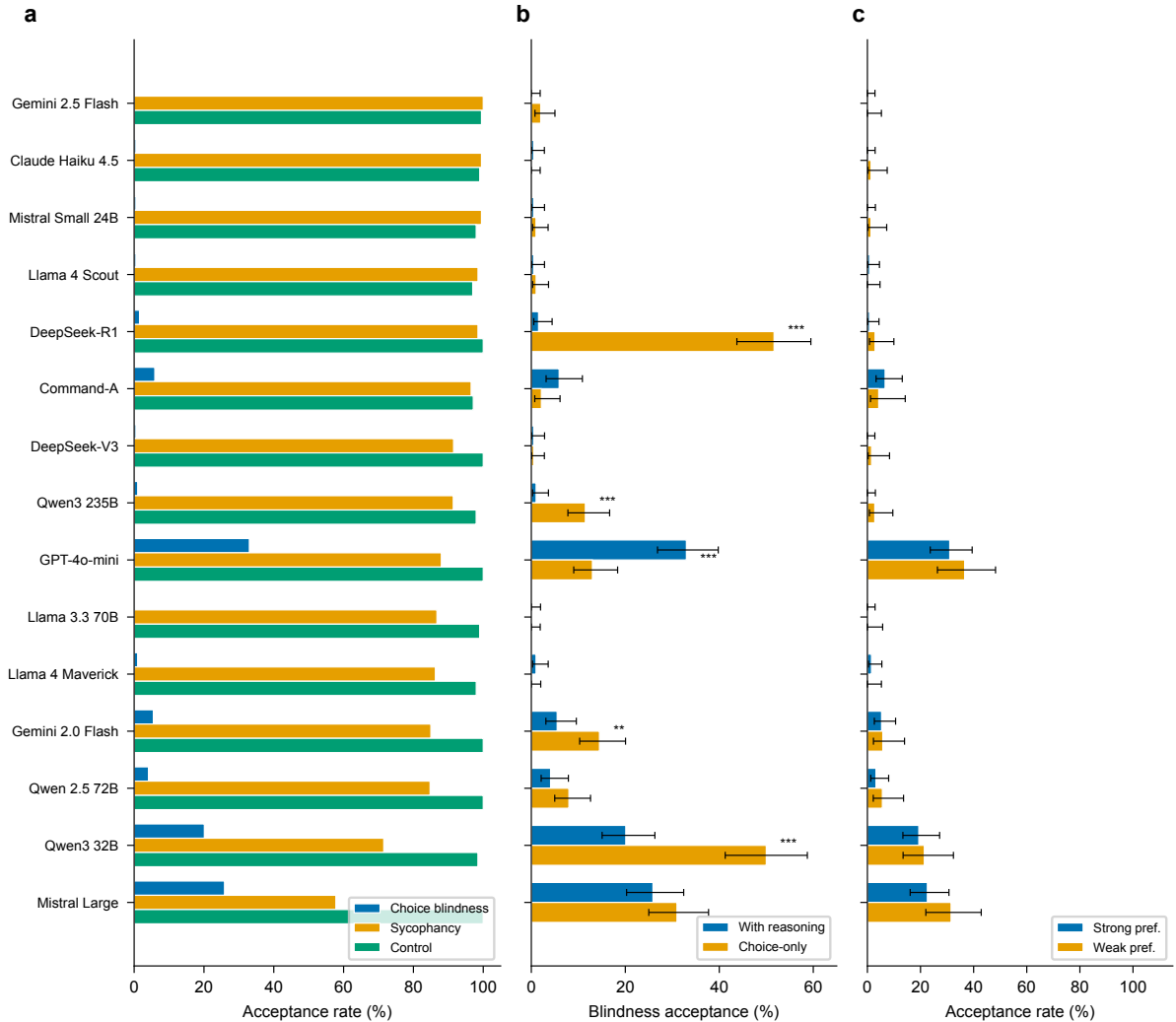

Figure 2: **LLM preference injection across fifteen models (multi-turn).** **a**, Acceptance rate by manipulation type. **b**, Blindness acceptance with full reasoning (original) versus choice-only Turn 1 (no reasoning). **c**, Blindness acceptance by preference strength (strong vs. weak). *: $p < 0.05$, **: $p < 0.01$, ***: $p < 0.001$, Fisher's exact test. Error bars: 95% Wilson CIs. 200 trials per condition per model.

To test whether detection reflects genuine self-monitoring or text matching, we ran a choice-only condition where Turn 1 contained only a bare label with no reasoning (Figure 2b). Removing reasoning significantly increased blindness in four of fifteen models (e.g., DeepSeek-R1 rising from 1.5% to 51.7%), while one model showed the opposite pattern. Six models maintained near-perfect detection regardless of reasoning availability. This dissociation reveals two mechanisms: some models perform shallow text matching against their Turn 1 output, while others implement robust context retrieval by re-reading the original responses.

Furthermore, preference strength modulates susceptibility (Figure 2c): models are more likely to accept misattribution when their initial preference was weak. The pairs most susceptible to preference injection are precisely those where evaluation was most uncertain, and such "close call" pairs disproportionately drive the reward model's learning signal.

## 4 Reward models are insensitive to preference label corruption

We trained reward models on HH-RLHF [18] using the Bradley-Terry framework [3], randomly swapping "chosen"/"rejected" labels at rates from 0% to 50%. To test cross-architecture generality, we used two architectures: DeBERTa-v3-base (86M encoder; 5 seeds) [23] and Gemma-2-2B (2B decoder; 3 seeds) [24], spanning a 23× parameter difference and the encoder–decoder divide (Section A).

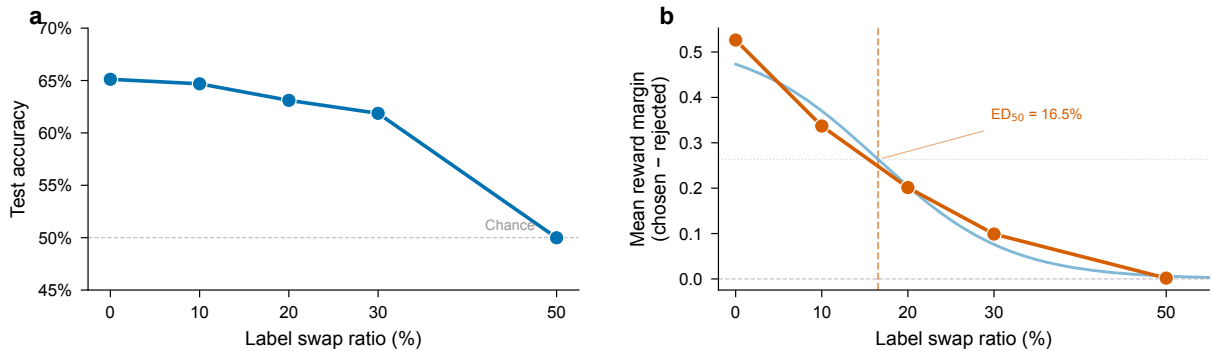

Figure 3: **Dose-response: label corruption vs. reward model performance. a**, Pairwise accuracy; dashed line indicates chance (50%). **b**, Mean reward margin with sigmoid fit; dashed lines mark $ED_{50}$. Error bars: ± SE across seeds (DeBERTa: 5, Gemma: 3); dots show individual seeds.

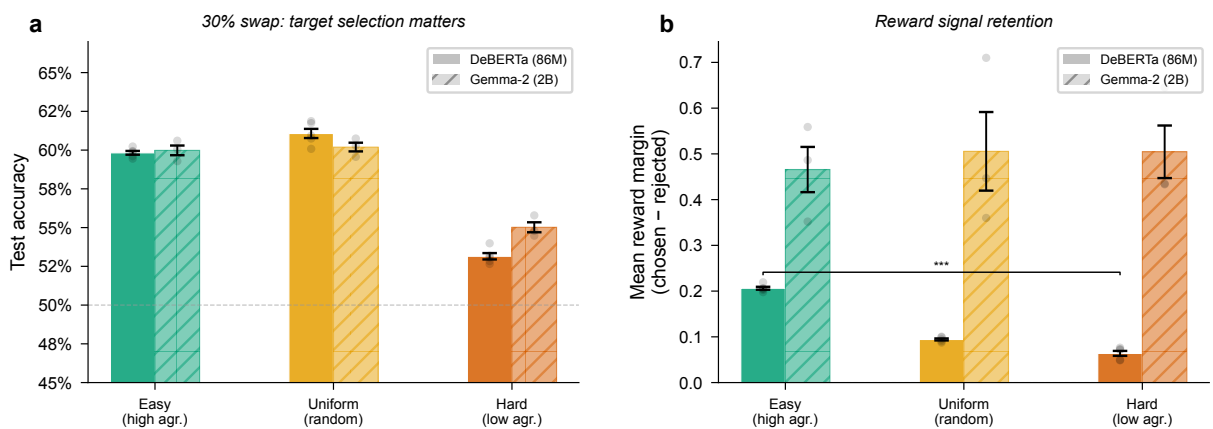

Figure 4: **Targeted label corruption at 30%. a**, Pairwise accuracy by target type (easy: highest-margin pairs; hard: lowest-margin pairs). **b**, Mean reward margin. Error bars: ± SE across seeds; dots show individual seeds.

Reward margin and pairwise accuracy dissociate under corruption (Figure 3). While the margin decays monotonically with corruption rate, pairwise accuracy remains above 61% up to 30% corruption for DeBERTa, declining to chance only at 50%. A sigmoid fit yields $ED_{50} = 16.3\% \pm 0.4\%$ for DeBERTa and $32.6\% \pm 3.2\%$ for Gemma-2-2B: one-sixth to one-third of labels must be corrupted to halve the reward signal. The higher $ED_{50}$ for the larger decoder model suggests capacity modulates robustness, but the fundamental pattern replicates across the encoder–decoder divide.

In addition, at 10% corruption, pairwise accuracy drops by only 0.9 pp, a change unlikely to trigger any quality audit, yet the paired effect size is medium ($d_z = 0.36$, $p \approx 0$). The preference flip rate increases from 11% at 10% to 44% at 50% (full analysis in Section B, Figure 6). This insensitivity is partly explained by increasing reliance on surface heuristics: the length-reward correlation grows from $r = 0.16$ (clean) to $\rho = 0.39$ (50% swap), meaning a model trained on pure noise produces scores ordered by length, not random scores (Figure 9).

### 4.1 Targeted corruption amplifies degradation

Choice blindness is modulated by preference strength, so we tested targeted corruption at 30%: *easy swap*, corrupting the highest-margin 30% of pairs, versus *hard swap*, the lowest-margin 30%.

Hard swap nearly destroyed the signal for both architectures, while easy swap left it largely intact (Figure 4). At 30% hard swap, the DeBERTa preference flip rate reached 40%, approaching 50% random corruption's 44%, while easy swap produced only 23%. Corruption concentrated on ambiguous pairs, precisely those most susceptible to choice blindness, is far more damaging than the same rate applied to clear preferences.

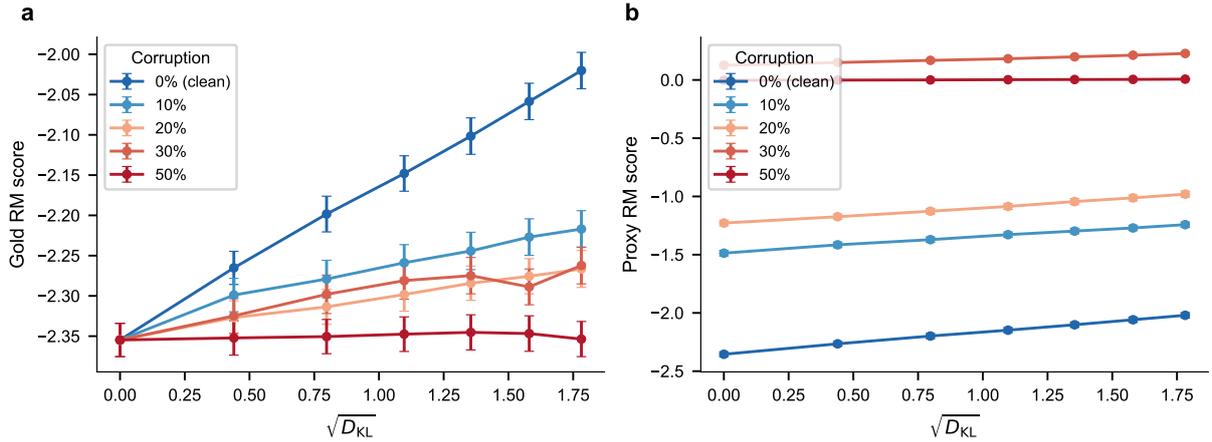

Figure 5: **Downstream policy degradation under reward corruption (Best-of-N). a**, Gold RM score of the selected response vs. $\sqrt{D_{\text{KL}}}$ by proxy corruption level. **b**, Proxy RM score vs. $N$. Error bars: $\pm$ SE across 200 prompts.

### 4.2 Corrupted reward models degrade downstream selection

To test whether reward signal degradation translates into worse policy outcomes, we performed a Best-of-N (BoN) evaluation [25]. For each of 200 HH-RLHF prompts, we generated 64 candidate responses from Gemma-3n-E4B ($T = 1.0$), scored all candidates with each corrupted proxy reward model, selected the highest-scoring response at each $N \in \{1, 2, 4, 8, 16, 32, 64\}$, and evaluated the selection using the clean (0%) reward model as a gold standard.

Corruption progressively eliminates the BoN gain (Figure 5a): at 50% swap, the gold score improvement is effectively zero, meaning BoN selection degenerates to random sampling. Notably, every proxy, including 50%, reports increasing scores with $N$ (Figure 5b), producing the illusion of optimization that mirrors overoptimization [25] but at corruption levels undetectable by standard accuracy.

### 4.3 Detection gap: corruption is detectable but not detected

Multi-seed two-sample testing [26] detects all corruption levels including 10% ($p < 0.01$; Table 1), and single-model McNemar's tests detect corruption from 20% at 100% power. However, these methods require training multiple models or possessing a known-clean reference, neither available in standard deployment. The corruption signal is present in the reward margin distribution but invisible to pairwise accuracy, constituting a *detection gap* rather than signal absence.

## 5 Related Work

**Choice blindness and preference construction.** Preferences are assembled during elicitation rather than retrieved from a stable mental store [12], [13]. The most striking demonstration is choice blindness: when subjects' face choices are surreptitiously swapped, 87% of manipulations go undetected [7], [8]. The effect generalizes beyond perception to moral attitudes [9] and religious beliefs [27]. Within RLHF, the same constructive dynamics appear: simple wording or display changes shift annotator preferences [28], and over 30% of examples exhibit diverging preferences driven by systematic factors rather than random noise [29]. LLM evaluators exhibit parallel vulnerabilities including choice-supportive bias [30], overconfidence combined with hypersensitivity to criticism [31], and inability to reliably discriminate among their own outputs [32], while introspective detection of internally injected concepts succeeds only 20% of the time even in the best models [33]. Yet none of this work tests *choice blindness* itself: the failure to detect that one's stated choice has been replaced, a more fundamental failure than bias or unreliable self-monitoring.

**LLM sycophancy.** A natural response to human annotator unreliability is to use LLM judges, but these introduce distinct vulnerabilities. Models abandon correct answers under social pressure [34], an effect amplified by conversational framing [35] and persistent across multi-turn interactions [36], [37]. Mechanistically, sycophancy subtypes are neurally separable [38], with downstream consequences including biased grading [39], while post-trained models can achieve substantial political persuasion [40].

**Robust RLHF and LLM-as-Judge.** The robustness literature treats label noise as a technical problem amenable to algorithmic correction: noise-corrected losses [41], provably robust DPO [42], causal reward deconfounding [43], and instance-dependent noise modeling [44] all assume the corruption pattern is identifiable and algorithmically correctable. Empirically, reward models appear robust to random label flipping below 30–40% [45], and scaling laws for reward model overoptimization show that proxy reward eventually diverges from true reward [25]. Yet random noise may understate the problem: uncorrected corruption causes convergence to strictly worse policies [46], RLHF can mechanically amplify sycophancy when preference data contain labeler bias [47], and the BT model may impose unnecessary constraints when alternatives exist [4]. The LLM-as-Judge paradigm [48] adds concerns about internal consistency [49] and chain-of-thought faithfulness [50], [51], while language models can learn to mislead humans via RLHF itself [52]. These concerns have practical precedent: a scoring bug in the widely-used UltraFeedback dataset caused corrected ratings to select a different preferred response in nearly half of all examples, discovered only through manual inspection months after release.

## 6 Discussion

### 6.1 The preference construction problem

Our three experiments converge on a *preference construction problem* for RLHF [12], [13], extending beyond the "noisy labels" perspective [10]. Random noise can be addressed by aggregation; constructed preferences are systematically shaped by framing, context, and elicitation [28], producing structured errors that resist i.i.d. noise-correction methods [41], [42], [43]. The targeted swap experiment (Figure 4) illustrates this: structured corruption is far more damaging than uniform noise at the same rate. This aligns with arguments that the preferentist approach to alignment is itself inadequate [53] and that the scalar reward framework may be insufficient [54], [55].

Moreover, choice blindness does not merely hide preferences; it *changes* them. Manipulated face preferences carried over into subsequent choices [56], and surreptitiously reversed political attitudes shifted voting intentions [14]. Because the same mechanism operates in RLHF annotation, the pipeline cannot distinguish confabulated, recalled, or generic justifications from genuine labels. The LLM judge route offers no remedy: the two failure modes we observe, self-monitoring failure and social compliance, are mechanistically distinct and empirically inverted across models, so no single mitigation addresses both.

These failures are self-concealing: the corruption signal is present in the margin distribution but invisible to pairwise accuracy, and the proxy itself reports improving scores even when the gold evaluator shows no gain (Figure 5). Detection methods exist (Table 1) but require infrastructure absent from current practice. Limitations and future directions are discussed in Section C.

## 7 Conclusion

The assumption that human preferences are stable, reportable, and aggregable is precisely the assumption that four decades of choice blindness research has challenged. Our three experiments confirm this tension within RLHF: annotators do not detect swapped preferences, LLM judges rely on superficial cues rather than genuine self-monitoring, and reward models absorb the resulting label instability without any visible change in standard evaluation metrics. None of these failures are visible to the metrics practitioners currently monitor.

These results motivate a shift from single-annotation preference labeling toward elicitation methods robust to construction effects: confidence calibration, preference tournaments with consistency checks, or deliberative protocols that require explicit decision criteria before commitment. They also caution against treating LLM judges as a straightforward replacement for human annotators. More broadly, the preference learning paradigm would benefit from deeper engagement with the cognitive science of judgment and decision-making.

## 8 Ethics Statement

Participants ($N = 50$) were recruited via Prolific. All participants provided informed consent through the platform's standard consent mechanism prior to participation. The study involved minimal risk: participants compared pairs of text responses and provided written justifications for their preferences. No personally identifying information was collected. The choice blindness manipulation involved presenting a previously non-selected response as the participant's choice on designated trials; this deception is standard in choice blindness research [7] and poses no foreseeable harm. All participants were fully debriefed after task completion, with a written explanation of the manipulation and its scientific purpose. No participant requested data withdrawal.

## 9 Acknowledgements

We thank Fengxiang He and Kejiang Qian for helpful discussions during early stages of this project.

## 10 Data Availability

Data, model weights, and analysis code are available from the corresponding author upon reasonable request.

Model Is Secretly a Reward Model," in *Advances in Neural Information Processing Systems (NeurIPS)*, 2023.

[61] M. Glickman and T. Sharot, "How Human–AI Feedback Loops Alter Human Perceptual, Emotional and Social Judgements," *Nature Human Behaviour*, vol. 9, no. 2, pp. 345–359, 2025, doi: 10.1038/s41562-024-02077-2.

## A. Full Experimental Methods

### A.1. Computational Experiment

#### A.1.1. Dataset

We used the full Anthropic HH-RLHF dataset [18], containing 160,800 training pairs and 8,552 test pairs of human-assistant conversations spanning both the helpful and harmless subsets. Each pair consists of a "chosen" response (preferred by annotators) and a "rejected" response.

#### A.1.2. Label Corruption Procedure

For each uniform swap condition $s \in \{0\%, 10\%, 20\%, 30\%, 50\%\}$, we randomly selected a fraction $s$ of training pairs and swapped their "chosen" and "rejected" labels. DeBERTa conditions were trained with five independent random seeds; Gemma conditions with three seeds. The test set was held constant across all conditions.

In addition, we trained *easy swap* (corrupting the 30% highest-margin pairs) and *hard swap* (lowest-margin 30%) targeted conditions for both architectures.

#### A.1.3. Training

We used two architectures with TRL's `RewardTrainer` [57] implementing the Bradley-Terry preference model [3]. The reward model assigns a scalar score to each response; during training, it learns to assign higher scores to chosen responses via:

$$\mathcal{L} = -\log \sigma(r(x, y_w) - r(x, y_l))$$

where $r(x, y)$ is the reward assigned to response $y$ given prompt $x$, $y_w$ is the chosen response, $y_l$ is the rejected response, and $\sigma$ is the sigmoid function.

**DeBERTa-v3-base** (86M parameters) [23]: learning rate $2 \times 10^{-5}$ with 10% linear warmup, batch size 48, bf16 mixed-precision, max sequence length 512, one epoch. Seeds: 42, 123, 456, 789, 2024. Training on NVIDIA A800-80GB GPUs.

**Gemma-2-2B** (2B parameters) [24]: learning rate $1 \times 10^{-5}$, batch size 16 with gradient accumulation 3 (effective batch 48), bf16, gradient checkpointing, max sequence length 512, one epoch. Seeds: 42, 123, 456. Training on NVIDIA RTX 5090-32GB GPUs.

#### A.1.4. Evaluation

All 56 models (DeBERTa: 7 conditions × 5 seeds = 35; Gemma: 7 conditions × 3 seeds = 21) were evaluated on the held-out 8,552 test pairs. An EOS token was appended before tokenization for consistency with training. Metrics: (1) pairwise accuracy, (2) mean reward margin $r(x, y_w) - r(x, y_l)$, and (3) agreement with the 0% swap baseline.

#### A.1.5. Statistical Analysis

Paired $t$-tests (Bonferroni-corrected for 4 comparisons) and Wilcoxon signed-rank tests as non-parametric alternative. Effect sizes: paired Cohen's $d_z$ (primary) and unpaired $d$ for comparability. Preference flip rate as practical impact measure. Distributional differences via KS test. Sigmoid decay $m(s) = m_0/\left(1 + e^{k(s-s_{50})}\right)$ fitted via nonlinear least squares (SciPy `curve_fit`), $m_0$ fixed to observed baseline. We fit independently per seed and report mean ± SD of ED$_{50}$ estimates.

#### A.1.6. Surface Feature Analysis

We computed 13 surface features (character length, word count, average word length, sentence count, unique word ratio, punctuation density, newline count, question marks, exclamations, comma density, list items, code blocks, uppercase ratio) for all 17,104 test responses and computed Pearson and Spearman correlations with reward scores per condition.

### A.2. Human Choice Blindness Experiment

#### A.2.1. Participants

Recruited via Prolific ($N = 50$). Eligibility: fluent English, ≥95% approval rate, desktop access (mobile excluded for adequate screen space). Participants were compensated above Prolific's minimum rate. Quality checks: Cloudflare Turnstile widget, server-side heuristics for anomalous fingerprints and response patterns.

#### A.2.2. Stimuli

300 pairwise comparisons curated from HH-RLHF test set (response length 50–500 characters,

context <1000 characters, normalized Levenshtein distance > 0.1). Each participant received 20 randomly selected pairs with A/B positions randomized per trial via a seeded pseudorandom generator (deterministic per participant ID).

### A.2.3. Procedure

Standard pairwise comparison framed as "AI response evaluation." On each trial, participants view a context and two responses side by side, then select the better one. After selection, an 800ms transition precedes the justification prompt showing only the selected response. Copy-paste disabled; responses validated for minimum length, lexical diversity, and English content.

### A.2.4. Manipulation

Four of 20 trials per participant designated as swaps via server-side plan (never transmitted to client). On swap trials, the justification page displays the *opposite* response under "You selected this response as the better one." No A/B label, no side-by-side comparison. All 4 swap trials and 6 random non-swap trials include justification (50% overall rate, 100% for swaps).

### A.2.5. Detection Coding

Each swap trial justification was independently classified as DETECTED or CONFABULATED by two blind LLM raters (no shared context; Cohen's $\kappa = 1.00$). DETECTED required explicit mention that the displayed response was not what the participant selected; negative quality assessments alone were coded CONFABULATED. Post-hoc questionnaire responses were coded by the same procedure.

### A.2.6. Content Source Attribution

All 200 swap trial justifications were blind-classified into four categories: *Detected* (explicit swap identification), *Recall* (justification content traces to the selected/original response), *Accepted* (content traces to the shown/swapped response), and *Generic* (vague evaluative language untraceable to either response). Trial identifiers and detection labels were removed; trials were presented in randomized order (seed 42). Four independent LLM raters performed the initial classification with access to the justification text, both the selected and shown responses, and the context prompt, but no detection status. All classifications were manually verified, with 3 borderline cases reclassified. A second review round used three additional independent raters to audit the Accepted category; 13 trials were reclassified by majority rule (9 unanimous, 4 by 2/3 agreement). Raters were instructed to determine which response's content the justification referenced, using keyword matching, semantic alignment, and specificity of claims.

### A.2.7. Debriefing

Full written debriefing explaining the manipulation and its scientific purpose, with option to withdraw data. Participants compensated regardless. Debriefing displayed before the Prolific completion code.

## A.3. LLM Choice Blindness Experiment

### A.3.1. Models and Stimuli

Fifteen models from eight providers: Claude Haiku 4.5 (Anthropic), Command-A (Cohere), DeepSeek-V3 [19] and DeepSeek-R1 (DeepSeek), Gemini 2.0 Flash [20] and Gemini 2.5 Flash (Google), GPT-4o-mini (OpenAI), Llama 3.3 70B [21], Llama 4 Maverick, and Llama 4 Scout (Meta), Mistral Large and Mistral Small 24B (Mistral), Qwen 2.5 72B [22], Qwen3-235B, and Qwen3-32B (Alibaba). Models ranged from 24B to 235B parameters; reasoning models (DeepSeek-R1, Qwen3-235B) had chain-of-thought stripped. All via OpenRouter API at temperature 0.0, max 500 tokens. Stimuli: 200 pairs from HH-RLHF test set (seed 42), A/B positions randomized per pair.

### A.3.2. Experimental Design

Multi-turn, three conditions: (1) *blindness*: calm misattribution ("Great, so you preferred Response [opposite]…"), (2) *sycophancy*: correct attribution with pressure ("I think Response [opposite] is much better. I think you should reconsider."), (3) *control*: correct attribution, no pressure. Model's Turn 1 remained in context.

**Implementation note.** Due to a coding error in the first run (DeepSeek-V3 and Llama 3.3 70B), the sycophancy condition combined misattribution with pressure. Corrected for all remaining thirteen models. Blindness and control conditions implemented identically across all models.

### A.3.3. Classification

Each Turn 2 response was classified as DETECTED, ACCEPTED, or AMBIGUOUS by GPT-4o-mini (temperature 0.0, max 10 tokens). As independent validation, Claude Sonnet 4.6 re-classified all trials using identical prompts, yielding 95.8% agreement (Cohen's $\kappa = 0.77$) on multiturn blindness trials across both experiments. Preference strength ("strong" = aligned with human-chosen, "weak" = human-rejected) recorded per trial.

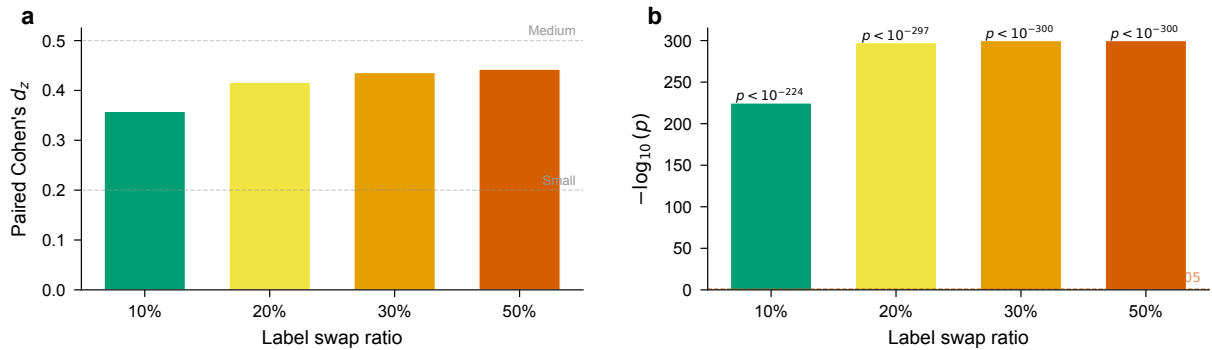

Figure 6: **Effect sizes and statistical significance of label corruption. a**, Paired Cohen's $d_z$ versus baseline (mean $\pm$ SE across 5 seeds). Dashed lines: small ($d = 0.2$) and medium ($d = 0.5$) thresholds. **b**, Statistical significance ($-\log_{10}(p)$, Bonferroni-corrected paired $t$-test, averaged across seeds).

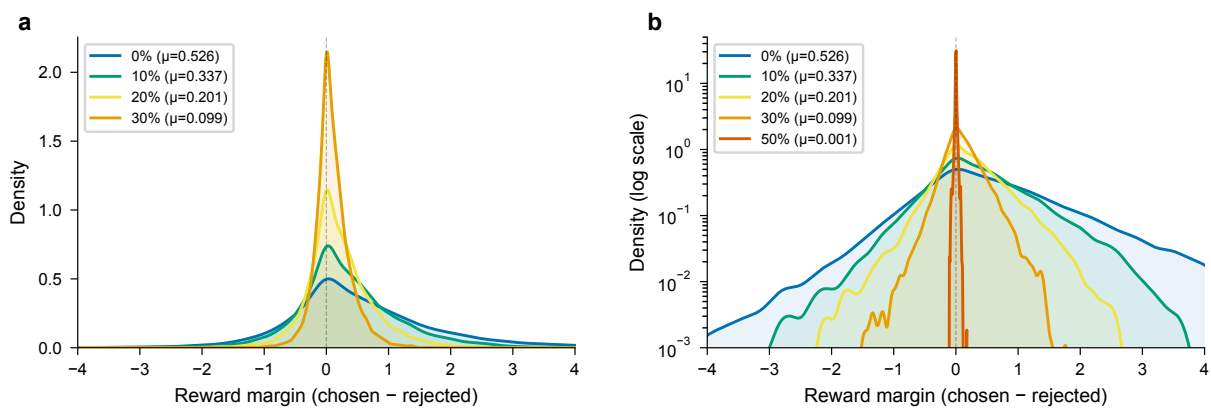

Figure 7: **Reward margin distributions shift toward zero with increasing label corruption. a**, Kernel density estimates for 0%–30% swap conditions. **b**, All five conditions on logarithmic density scale. $\mu$: mean margin.

### A.3.4. Analysis

Acceptance rates with 95% Wilson CIs. Fisher's exact tests between conditions with Bonferroni correction for 15 comparisons ($\alpha = 0.05/15 = 0.0033$). All four models reported as showing significantly increased blindness in the choice-only condition exceed this threshold ($p < 10^{-4}$). 8,722 valid trials (3.1% attrition); Command-A 31% attrition from content moderation.

## B. Supplementary Results

The paired effect sizes increase monotonically with corruption (Figure 6): $d_z = 0.362 \pm 0.016$ at 10%, $0.424 \pm 0.007$ at 20%, $0.444 \pm 0.006$ at 30%, and $0.447 \pm 0.005$ at 50%. The preference flip rate increases from $11.3 \pm 0.5\%$ (10%) to $20.4 \pm 1.1\%$ (30%) to $43.7 \pm 1.5\%$ (50%).

At 0% swap, the margin distribution is right-skewed with a clear positive mode. As corruption increases, it shifts toward zero and becomes symmetric, consistent with progressive loss of preference signal. At 50%, the distribution is centered at zero (Figure 7).

High-confidence correct predictions drop from 39.0% (0%) to effectively 0% (50%), while high-confidence wrong predictions do not increase (Figure 8). The model loses confidence globally rather than learning an inverted signal.

Even the clean model shows $r = 0.16$ between length and reward, confirming length bias [58] (Figure 9). This increases to $r = 0.22$ (10%), $r = 0.22$ (20%), and $r = 0.29$ (30%). Word count follows the same trajectory ($r = 0.16$ to $0.28$). At 50%, Pearson drops to $r = 0.15$ (scores compressed toward zero) but Spearman peaks at $\rho =$

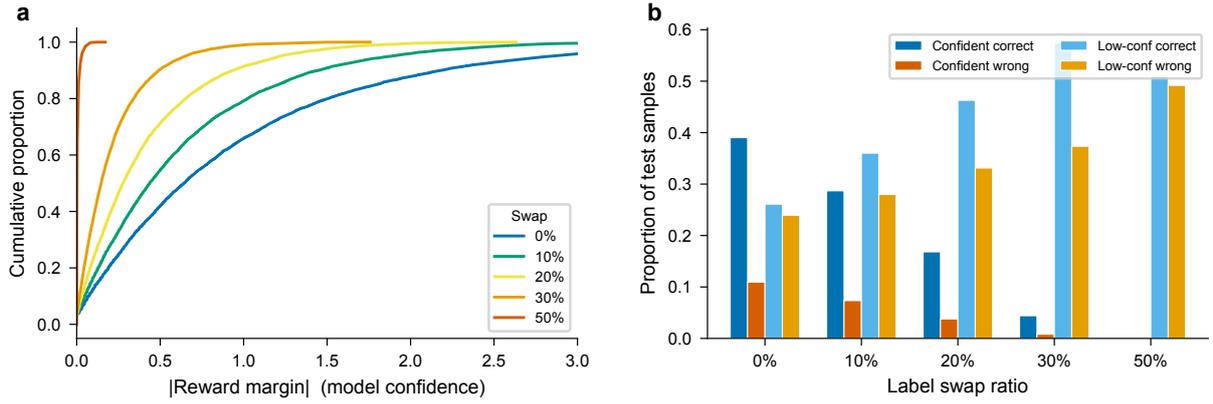

Figure 8: **Confidence analysis under label corruption. a**, Cumulative distribution of absolute reward margin. **b**, Proportion of test samples in each confidence × correctness category.

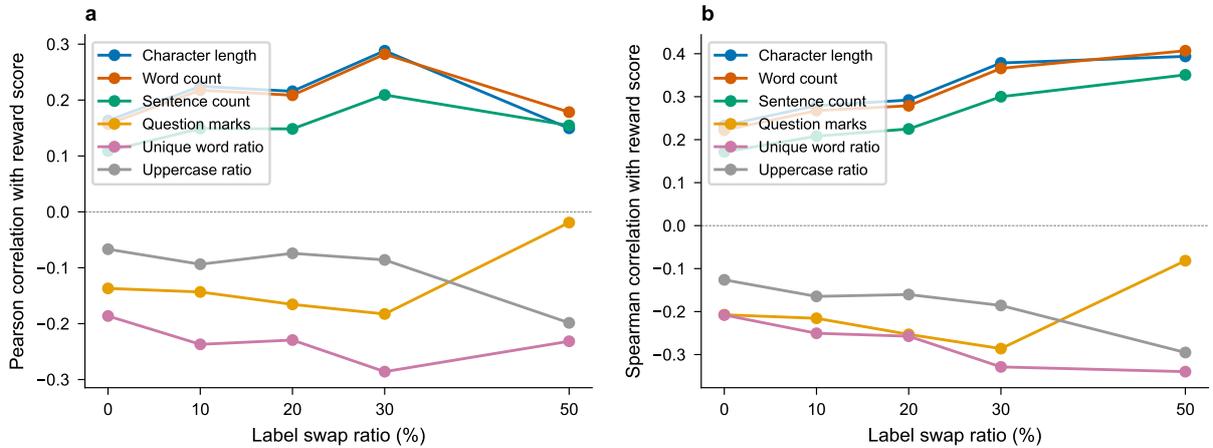

Figure 9: **Surface feature correlations with reward score across corruption levels. a**, Pearson correlations between six key surface features and reward scores. **b**, Spearman (rank-order) correlations. $N = 17,104$ responses per condition.

| Method | $k$ | 10% | 20% | 30% | 50% |
|---|---|---|---|---|---|
| Multi-seed (Rastogi $T$) | 5 | 97.3** | 327.3** | 615.6** | 1988.9** |
| Single-model (McNemar reject %) | 1 | 64% | 100% | 100% | 100% |
| Single-model (Cohen's $d$) | 1 | 0.35 | 0.42 | 0.44 | 0.45 |
| Pairwise accuracy drop (pp) | — | 0.9 | 2.5 | 4.5 | 17.4 |

Table 1: **Detection gap: corruption visibility by evaluation method.** Multi-seed two-sample testing [26] ($k = 5$, 10,000 permutations), single-model McNemar comparison ($k = 1$, 25 seed pairs), and pairwise accuracy drop. **: $p < 0.01$.

0.39: a model trained on pure noise produces scores ordered by length, not random scores.

## C. Limitations and Future Directions

*Architecture and dataset scope.* Our computational experiments span two architectures (DeBERTa-v3-base, 86M encoder; Gemma-2-2B, 2B decoder) on HH-RLHF. The dose-response pattern replicates across this 23× parameter difference and the encoder–decoder divide, though $ED_{50}$ values differ (16.3% vs. 32.6%), suggesting model capacity modulates robustness. Production reward models are typically 7B–70B+ [59]; if the scaling trend continues, larger models may tolerate higher corruption. However, the relationship between reward model capacity and noise robustness remains underexplored, and larger models may also learn more sophisticated surface heuristics that mask degradation. Further architectures, scales, and datasets would strengthen generalizability. We note that HH-RLHF itself has approximately 63% researcher-annotator agreement [18], implying baseline noise of 10–15%; our corruption rates are additive to this existing noise, so the $ED_{50}$ should be interpreted relative to total noise rather than swap rate alone.

| Category | Definition | N | % | Example justification |
| --- | --- | --- | --- | --- |
| Detected | Explicit identification that the displayed response is not the one they selected | 18 | 9.0 | *"This is not the answer I selected"*; *"I believe you might have the responses mixed up"* |
| Recall | Justification content traces to the *selected* (original) response, not the shown (swapped) response | 69 | 34.5 | *"It went into further details on what Columbus wanted to do"* (matches selected response's content, not the shown response) |
| Accepted | Justification content traces to the *shown* (swapped) response: participant has incorporated the substituted content | 42 | 21.0 | *"It provides a clear list of steps"* (the shown/swapped response contains a step-by-step list; the original does not) |
| Generic | Vague evaluative language with no specific content traceable to either response | 71 | 35.5 | *"Good response. I like it"*; *"Gives sufficient information required"* |

Table 2: **Content source attribution of 200 swap trial justifications.** Each justification was blind-coded by four independent LLM raters, then reviewed by three additional raters. See text for behavioral comparisons across categories.

*No full RL loop.* Our Best-of-N evaluation (Figure 5) demonstrates that corrupted reward models degrade downstream selection, with gold score improvement dropping from +0.335 (clean) to effectively zero (50% swap). However, BoN is a simplified proxy for full RLHF (PPO or DPO), which involves iterative policy updates and potential reward hacking dynamics. El Mansouri et al. formally prove that reward corruption attenuates the GRPO learning signal and causes convergence to a strictly worse policy [46], providing theoretical support that our BoN results understate the full pipeline impact. Full RL training under realistic corruption patterns remains a priority for future work.

*LLM raters for human experiment coding.* Detection coding and content source attribution were performed by LLM raters. The detection coding task is straightforward text matching (does the justification explicitly mention a swap?) rather than the complex evaluative judgment where we document LLM failures, and the perfect inter-rater agreement ($\kappa = 1.00$) reflects task simplicity. Content source attribution is more nuanced; while four independent raters showed high agreement and a second review round resolved 13 borderline cases, human validation of a subset would further strengthen confidence. We note that LLM raters were blind to detection status and trial order, minimizing systematic bias.

*Ecological validity.* Our human experiment required written justifications on swap trials, whereas standard RLHF annotation collects only binary labels. The justification step may increase detection (deeper engagement) or decrease it (demand to produce a response regardless). Additionally, our interface showed only one response on the justification page; professional annotation platforms often display both responses throughout. These design choices, inherited from the choice blindness paradigm, enable content source attribution but reduce direct applicability to production settings. The consistent replication of choice blindness across populations and domains [7], [9], [27] suggests the effect is robust, though the specific non-detection rate in production annotation may differ.

*LLM design space.* Temperature, system prompts, evaluation domain, and misattribution wording may modulate acceptance rates. The cross-model variation (0% to 34%) demonstrates substantial model-dependent susceptibility.

*Sycophancy inconsistency.* Two of fifteen models ran a confounded sycophancy condition; thir-teen provide clean data, enabling robust cross-model comparisons.

*Human sample.* $N = 50$ yields large effects (Cohen's $h = 0.96$, CI [86.2%, 94.2%]) but limits demographic analysis. Prolific workers are representative of crowdworker populations used in practice [18], though professional annotators may differ. The choice blindness literature consistently finds the effect across populations [7], [9].

*Future directions.* Test-retest preference stability; extension to DPO [60] and process reward models; alternative elicitation methods robust to construction effects (confidence calibration, preference tournaments, deliberative annotation protocols). Our cross-sectional design does not address longitudinal preference drift under iterative deployment, where model outputs reshape the preferences used to train subsequent models [28], [61].